# pH Prediction by Artificial Neural Networks for the Drinking Water of the Distribution System of Hyderabad City

NIAZ AHMED MEMON*, MUKHTIAR ALI UNAR**, AND ABDUL KHALIQUE ANSARI***

## ABSTRACT

In this research, feedforwardANN (Artificial Neural Network) model is developed and validated for predicting the pH at 10 different locations of the distribution system of drinking water of Hyderabad city. The developed model is MLP (Multilayer Perceptron) with back propagation algorithm. The data for the training and testing of the model are collected through an experimental analysis on weekly basis in a routine examination for maintaining the quality of drinking water in the city. 17 parameters are taken into consideration including pH. These all parameters are taken as input variables for the model and then pH is predicted for 03 phases; raw water of river Indus, treated water in the treatment plants and then treated water in the distribution system of drinking water. The training and testing results of this model reveal that MLP neural networks are exceedingly extrapolative for predicting the pH of river water, untreated and treated water at all locations of the distribution system of drinking water of Hyderabad city. The optimum input and output weights are generated with minimum MSE (Mean Square Error) < 5%. Experimental, predicted and tested values of pH are plotted and the effectiveness of the model is determined by calculating the coefficient of correlation ($R^2=0.999$) of trained and tested results.

Keywords: ANNs, pH modeling, drinking water quality, distribution system.

## 1. INTRODUCTION

The drinking water if polluted by chemical, physical and biological contaminants all over the world can be considered as an epidemic problem [1-2]. Similarly, quality of river water at any point reflects several major influences including the lithology of the basin, atmospheric inputs, climatic conditions and anthropogenic inputs [3]. Besides that, river water plays a foremost task in assimilation or transporting municipal and industrial waste water and runoff from agricultural lands [4]. Therefore, river quality assessment is of great importance because it directly influences public health (via drinking water) and aquatic life (via raw water) [5]. River water pH is affected by numerous processes; geochemical, physical and biological. Interrelationship between these processes and pH values are complex, non-linear and not well understood [6]. Drinking water; being an essentiality; for human being needs specific and periodic monitoring with respect to its quality and aesthetic parameters including pH; being referred as one of the National

* Assistant Professor, Department of Civil Engineering, Quaid-e-Awam University of Engineering, Science & Technology, Nawabshah.
** Professor, Department of Computer Systems Engineering, Mehran University of Engineering & Technology, Jamshoro.
*** Adjunct Professor, Department of Chemical Engineering, Karachi University, Karachi.

Secondary Drinking Water standards [7-8]. The NOM (Natural Organic Matter) during disinfection interactions depend on solution chemistry parameters including ionic strength, pH and concentrations of multivalent ions [9-11]. If NF (Nanofiltration) control is applied during the municipal water treatment, it may comprise of a mixture of ions of different charges [12-13]. The pH of a solution is the negative common logarithm of the hydrogen ion activity and the pH of most drinking water lies within the range 6.5-8.5. Although pH usually has no direct impact on water consumer, it is one of the most important operational water quality parameters and it needs careful attention to be controlled at all stages of water treatment to ensure satisfactory water clarification and disinfection. For effective disinfection with chlorine, the pH should be <8.0 and it must be controlled while entering in the distribution system to minimize the corrosion of water mains and pipes in household water systems, failing to that contamination of drinking water and adverse effects on its taste, odor and appearance can occur [14]. It is usually measured electrometrically [15-17]. This study aims to observe and model the pH of drinking water at all locations starting from the river intake, during settlement and finally in the distribution system. Ascertaining the quality of drinking water while reducing operating costs, advanced process control and automation technologies are introduced in early years, specially the use of artificial neural networks increased in the drinking water treatment because of its robust nonlinear modeling of complex unit processes [18]. ANNs are information processing networks constituting a set of highly interconnected neurons arranged in multiple layers that can be trained to fit one or more dependent variables as inputs [19]. Once the ANN is trained using experimental data, it can be used in a purely predictive mode to calculate the dependent variable(s) for any values of input variables. Historically, these have been used successfully to predict membrane fouling during micro and ultra filtration colloids of drinking water system [20], proteins [21], as well as industrial [22-23] and municipal waters [24]. The ANN can lucratively predict the Electrical Conductivity in combination of the parameters (input variables) like turbidity, temperature, pH, and chlorine at different locations of distribution system of drinking water [25] and also can predict precise TDS (Total Dissolved Solids) in the water supply distribution system [26]. This study aims to develop MLP network with backpropagation algorithm to predict the pH at different locations of the distribution system of drinking water of Hyderabad city. The number of the input variables for this model is 17.

## 2. ARTIFICIAL NEURAL NETWORKS

The architecture of neural network developed in this study consists of one input layer, one hidden layer and one output layer, as shown in Fig. 1.

The backpropagation algorithm incorporating the Levenberg Marquardt algorithm is used to minimize the MSE during the training and learning process. The transfer function used in the hidden layer is "tansig", which calculates a layers output from its net input. The transfer function of the output layer is a linear function.

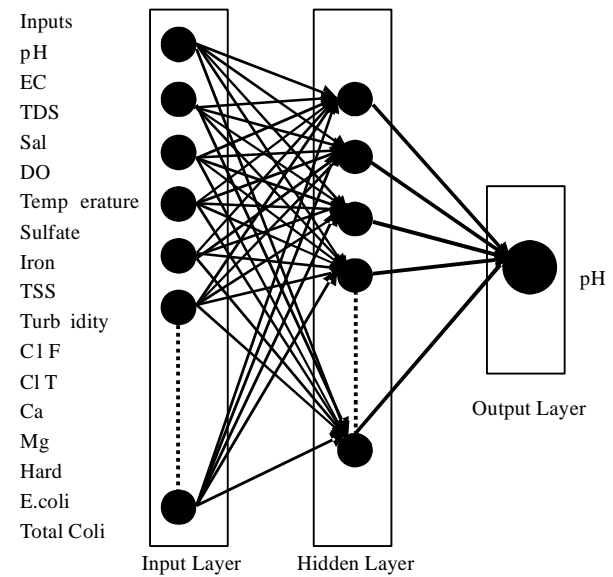

FIG. 1. MLP NETWORK

## 3. STUDY AREA: THE CITY OF HYDERABAD

The city of Hyderabad is one of the biggest cities of Pakistan with approximately 1.8 million population. The city lies with a latitude of 25° 22' North and a longitude 65° 41' with typical arid topography [27]. The main source for supply of the drinking water to this city is the River Indus. Two treatment plants are employed for supply of treated water to the dwellers of the city. The existing and proposed drinking water distribution system is shown in Fig. 2.

## 4. MATERIALS AND METHOD

Ten (10) different sampling locations are selected starting from the River intake (Location-1). The water samples are collected twice a week on these randomly selected locations. We have followed the WHO (World Health Organization) guidelines for selecting the locations and collecting the samples of the water. The sampling period started from the month of February-July 2009, covering winter and summer seasons. The HACH Spectrophotometer DR 2700 is used for testing of the iron (Fe), Chlorine total and Chlorine Free, Ca, Mg, Hardness,

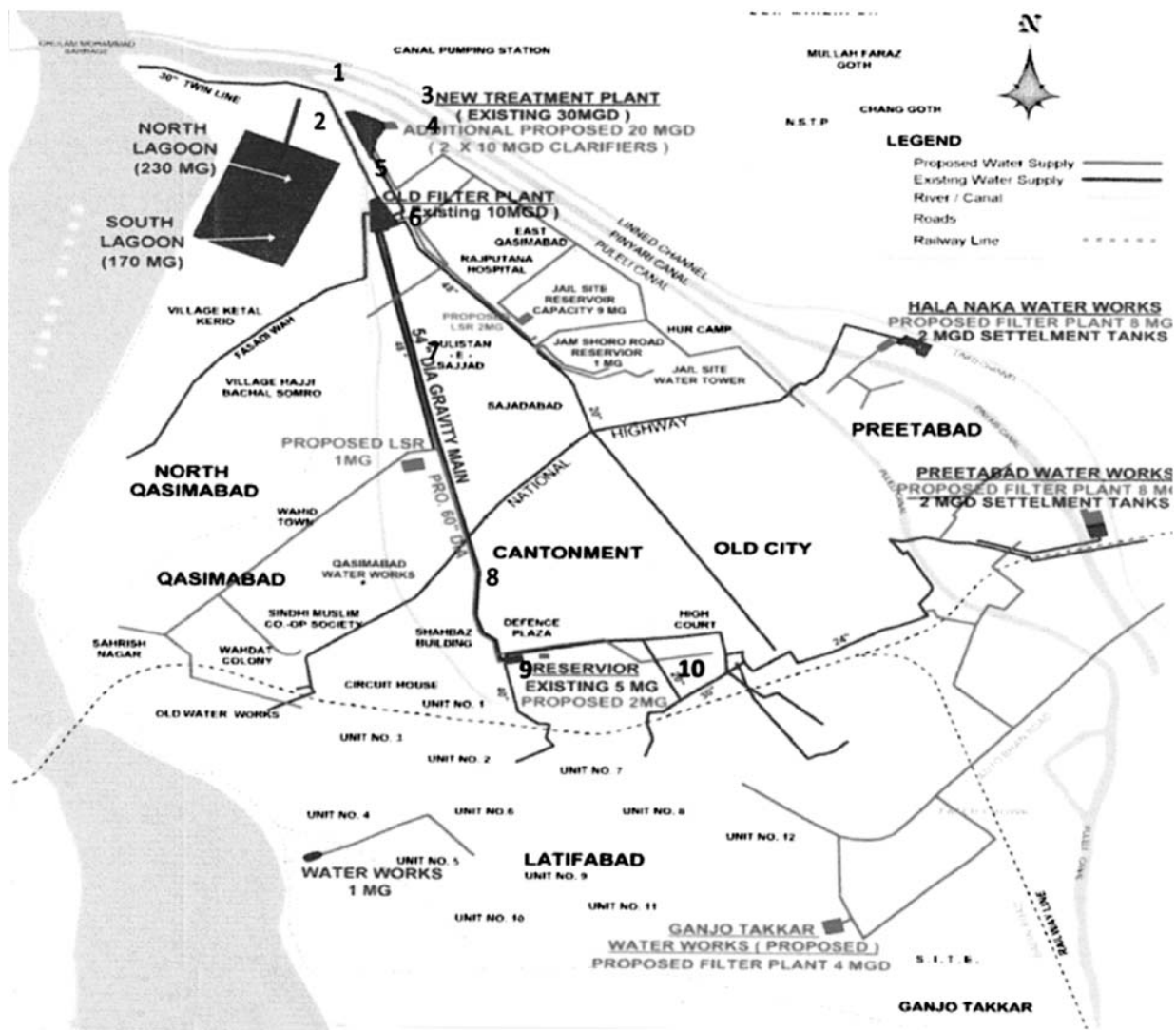

*FIG. 2. LOCATION MAP OF THE STUDY AREA*

Total suspended solids and sulfates and the turbidity is also tested by this equipment. The Sension 59 HACH Multiparameter meter is used for testing of pH, Electrical conductivity, Total Dissolved Solids, Salinity, Temperature, and Dissolved Oxygen. One Feedforward MLP ANN model with backpropagation algorithm is developed for calibration, and validation to predict the pH of rawwater, treated water in treatment plants and drinking water in the distribution system of Hyderabad city.

## 5. RESULTS AND DISCUSSION

### 5.1 Experimental Results

The experimentally observed pH results of raw water taken at the river intake, at Kotri Barrage, Jamshoro, location-1, are shown in Fig. 3. This is the location from where the raw water is pumped daily to the lagoons for storage and then pumped to the treatment plants. The observed values of the pHof settled water (Lagoon) are shown in Fig. 4. The mean value of pH at location-1 is 7.37(6.8-7.9). Similarly at location-2 the pH mean value is 7.4(6.9 - 7.81).There is a slight increase of pH in Lagoon water; however, these values lie within the guidelines available from WHO [28].

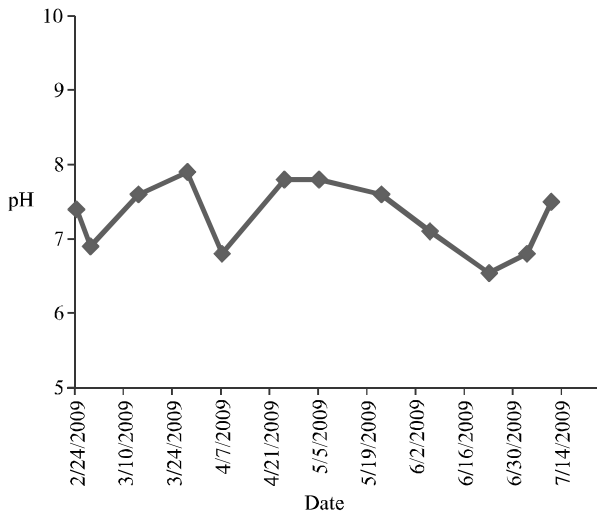

*FIG. 3. pH OBSERVED VALUES AT RIVER INDUS (LOCATION-1)*

The raw water is continually pumped to the NTP (New Treatment Plant) having the capability of 30 MGD supply of treated water to the city. The pH is observed at this location, the inlet of NTP, where raw water enters in the New Treatment Plant (location-3). Thesevalues are shown in Fig. 5. The pH values observed at the filtration gallery (outlet) of NTP (Location-4) are shown in Fig. 6. The pH mean value at location 3 is 7.3(6.9 -7.7) and the pH mean value at location-4 is 7.34(7.5-7.7). There is no significant change in pH values of raw water and treated water.

The treated water from NTP is directly supplied to a mixing chamber where treated water from OTP (Old Treatment Plant) is also received for onward supply to the city through different pumping stations in the distribution system. We have observed the pH values at the outlet of this mixing chamber located at OTP (Location-6) as shown in Fig. 6. The mean value is 7.4(7.1-7.7). No significant change in pH is observed during the flow from NTP to OTP.

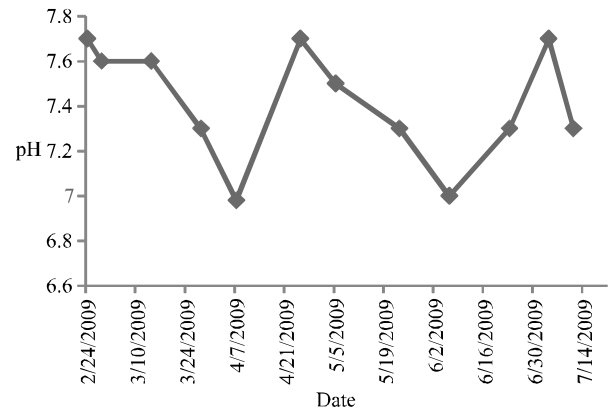

*FIG. 4. pH OBSERVED AT THE LAGOON (LOCATION-2)*

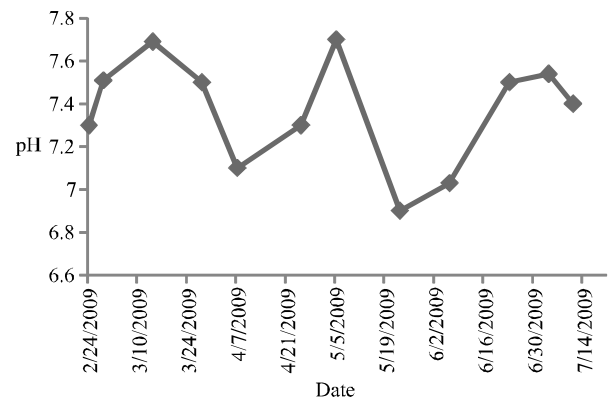

*FIG. 5. pH AT INLET OF NTP (LOCATION-3)*

In this study, two more locations are presented from the distribution system of the drinking water, to depict the quality of the drinking water throughout the whole system. The pH observation made at 1 MGD Pumping station, opposite to Citizen Colony, near Agha Khan Hospital, (Location-7) is shown in Fig. 7. The mean value is 7.54(7.1-7.85). An increase of 1 unit of pH as a mean value is increased between these two consecutive stations, which needs source identification.

The second location selected from the distribution system is the outlet of the LSR Pumping station (Location-10). The observed pH values of this location are shown in Fig. 8. The mean value is 7.37(7-7.9). Significant change in reduction of pH between these 02 stations is observed. However, all the values at all locations are consistent within WHO guidelines.

## 5.2 Modeling Results

MLP Neural network model is developed, trained and tested for predicting the pH at the locations discussed in Section 5.1. In this study 70% of the data are used for training sets and 30% as testing sets. The training graph of MLP is shown in Fig. 9. The MSE graph for training phase is presented in Fig. 10 and for testing, in Fig. 11.

In this Section, 3 model results of one location from each zone are presented. Location-1 represents the raw water, location-4, the treated water within treatment plant and location-7 represents the drinking water in the distribution system of the city. The results presented are enough to demonstrate the predictive potential of the trained ANN models.

The observed, trained and tested value of pH of raw water of Indus River at Kotri Barrage Jamshoro (Location-1) is shown in Fig. 12. The model fits the data well and expresses 89% of pH variance. The correlation coefficient is high in calibration set ($R^2=0.996$) as shown in Fig. 13; as well as in the verification set ($R^2=0.99$) as depicted in Fig. 14. The model conserves the same mean as the mean of the experimental data.

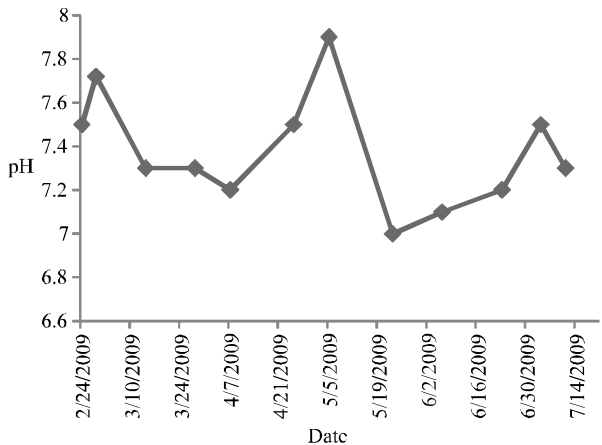

*FIG. 6. pH AT THE FILTRATION GALLERY OF NTP (LOCATION-4)*

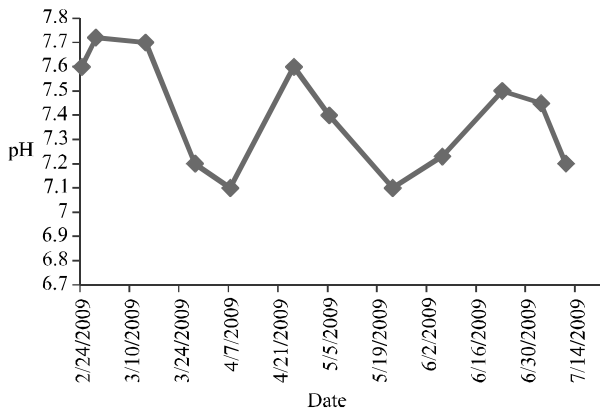

*FIG. 7. pH AT OUTLET OF MIXING CHAMBER (LOCATION-6)*

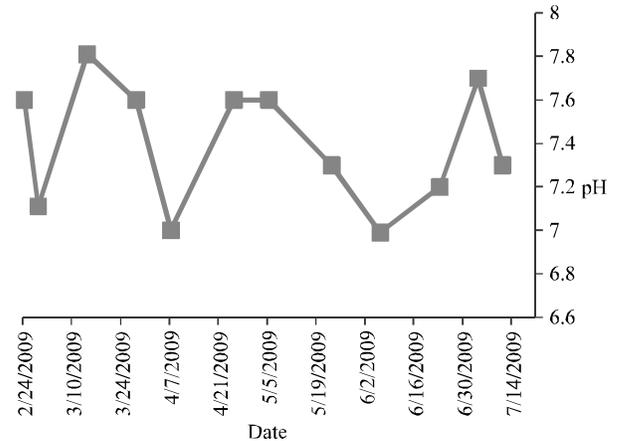

*FIG. 8. pH AT THE OUTLET OF LSR PS (LOCATION-10)*

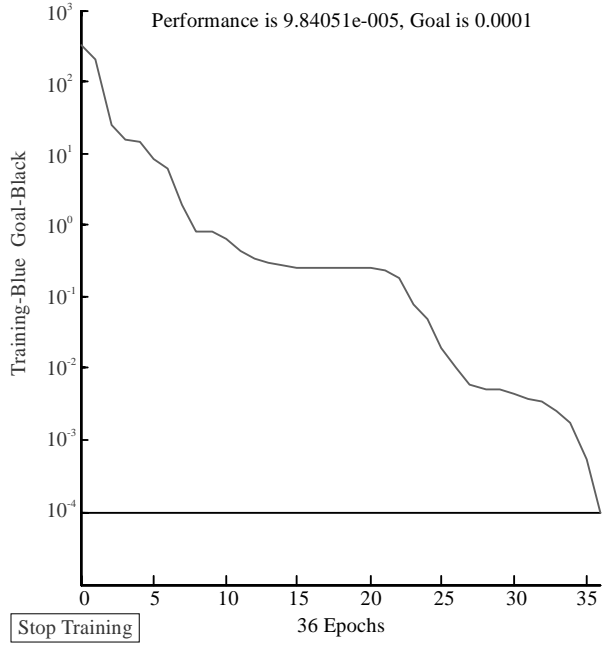

FIG. 9. MLP TRAINING

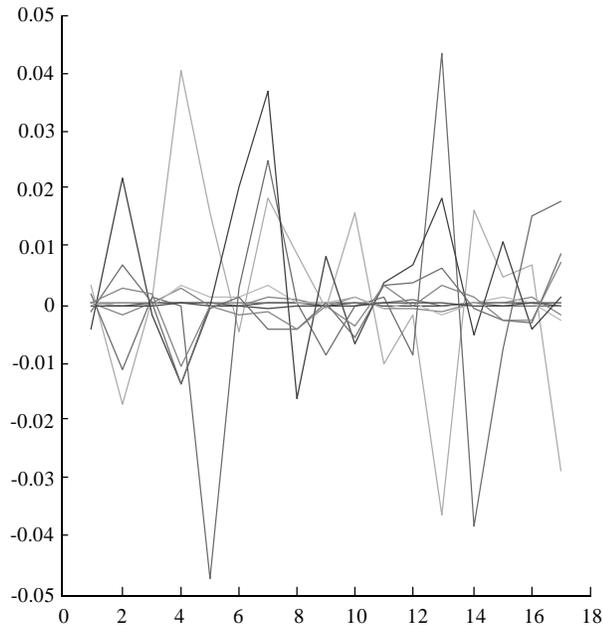

FIG. 10. MSE (TRAINING)

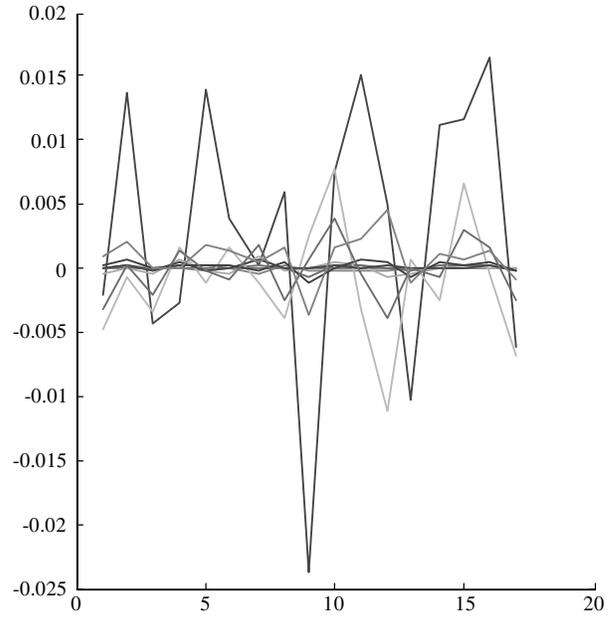

FIG. 11. MSE (TESTING)

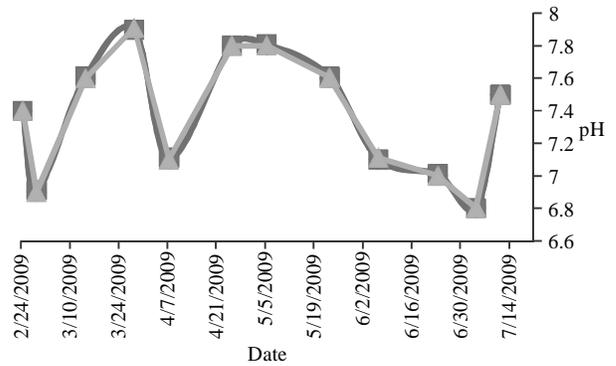

FIG. 12. OBSERVED, PREDICTED AND TESTED *pH* VALUES AT LOCATION-1

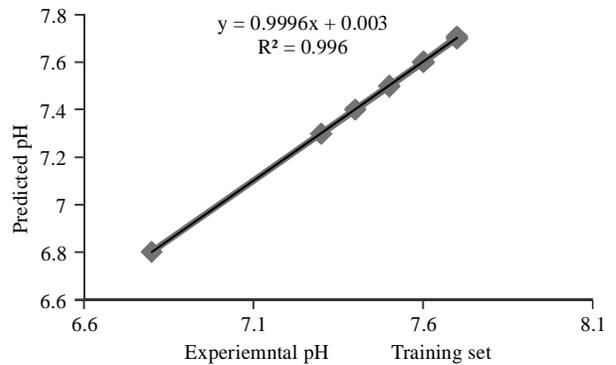

FIG. 13. PREDICTED VERSUS OBSERVED pH (LOCATION-1)

Our results presents more accuracy and are consistent with [6], who presented $R^2=0.88$ for calibration and $R^2=0.86$ for validation of ANNs model for predicting pH of raw water of Eutrophic Middle Loire River.

Similar trend of the predictive efficiency of the model is apparent; while calibrating and validating the pH of treated water within the treatment plants (Location-4) the observed, predicted and tested results of the model for this location are presented in Fig 15 and the corresponding correlations coefficient for calibration (training set) and validation (testing set) are presented in Figs. 16-17. The correlation represents the predictive competence of the model to predict this parameter of this study. The coefficient of correlation for training set is $R^2=0.8901$ and for testing set, $R^2=0.884$., the observed mean value of pH is consistent with the mean of the observed data at this location.

In this study the developed MLP model has shown its capability of forecasting the pH in the distribution system as well. The experimental, predicted and validated results of pH at the outlet of LSR Pumping station Thandisarak (Location-10) are shown in Fig. 18. The results are in agreement with [29], as the results obtained in that study demonstrated 6% MSE with a note that an ANN of MLP type is capable of forecasting daily values of salinity in the RiverMurray at Murray Bridge in South Australia. We have <5% MSE in this study. The correlation coefficient of calibrated and validated sets are ($R^2=0.882$ and $0.877$ respectively) as shown in Figs. 19-20.

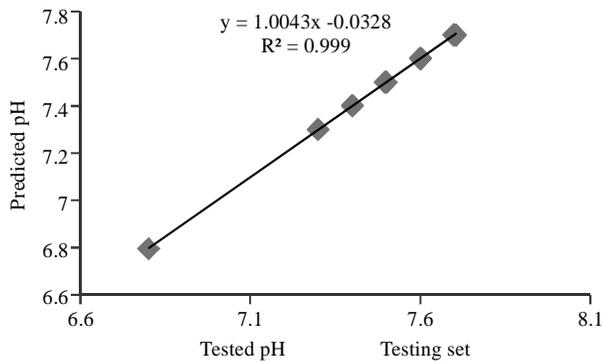

FIG. 14. PREDICTED VERSUS TESTED pH (LOCATION-1)

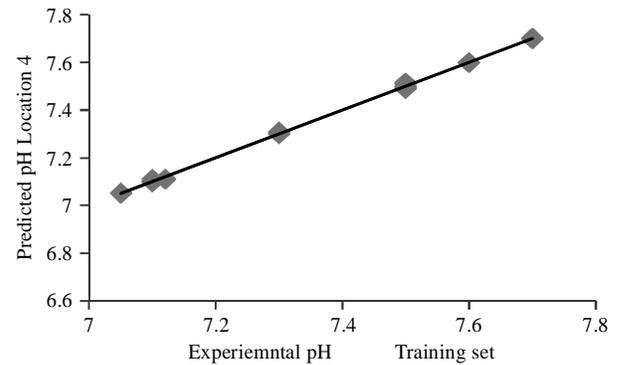

FIG. 16. PREDICTED VERSUS OBSERVED pH (LOCATION-4)

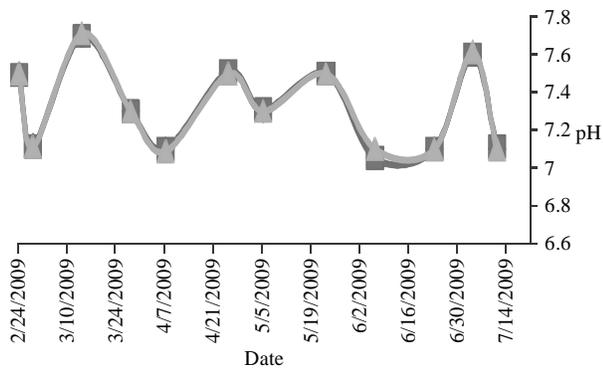

FIG. 15. OBSERVED, PREDICTED AND TESTED VALUE OF pH AT LOCATION-4

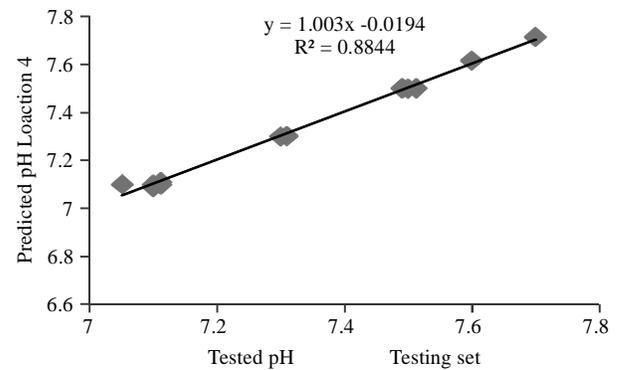

FIG. 17. PREDICTED VERSUS TESTED pH (LOCATION-4)

*pH Prediction by Artificial Neural Networks for the Drinking Water of the Distribution System of Hyderabad City*

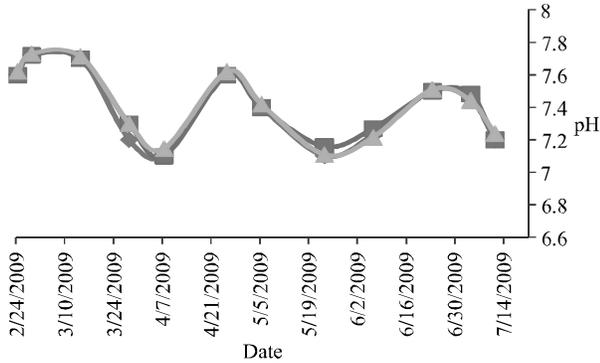

FIG. 18. OBSERVED, PREDICTED AND TESTED pH AT LOCATION-10

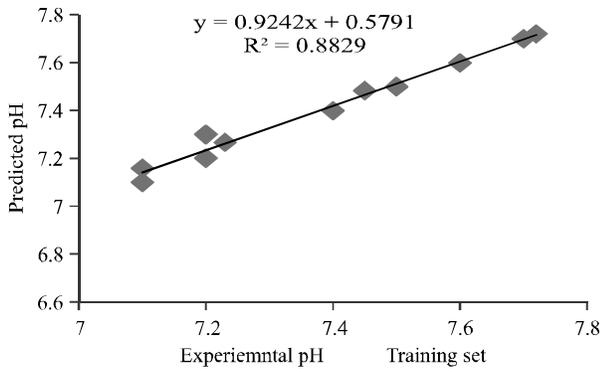

FIG. 19. PREDICTED VERSUS OBSERVED pH (LOCATION-4)

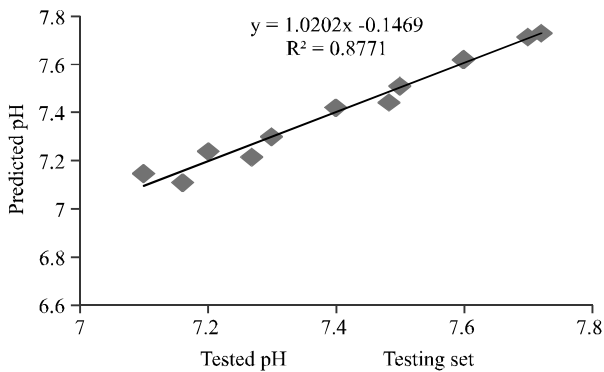

FIG. 20. PREDICTED VERSUS TESTED pH (LOCATION-4)

## 6. CONCLUSIONS

It is concluded from this study that the pH of river Indus at Kotri Barrage, Jamshoro lies within WHO guidelines with a slight increase in the lagoon water during storage. The pH values at 1 MGD pumping station increases slightly but it is consistent with WHO guidelines. The outcome of this study indicate that the Feedforward ANNs can give satisfactory response in modeling of pH in raw water of river Indus, treated water during the treatment process and drinking water in the distribution system as a function of drinking water quality data. The best coefficient of correlation for testing network is ($R^2$=0.999) found to predict the pH of river (raw water) at the intake location from where the water is taken for treatment and distribution to the city of Hyderabad. We contend that the similar results could be obtained for other critical parameters of the drinking water either during treatment or in the water supply system of the city.

## ACKNOWLEDGEMENTS

The authors acknowledge the facilities provided by the Mehran University of Engineering & Technology, Jamshoro, Pakistan, for conducting the tests and experimental work for this research in the laboratory of Environmental Engineering. The coordination and cooperation extended by the WASA, specially the authorities of New Treatment Plant (30 MGD), Hyderabad, Pakistan, for collecting samples, is highly appreciated.## REFERENCES

[1] Noor, R., and Karbbasi, A.R., "Prediction of Municipal Solid Waste Generation with Combination of Support Vector Machine and Principal Component Analysis: A Case Study of Mashhad", Journal of Environmental Progress & Sustainable Energy, Volume 28, pp. 249-258, 2009.

[2] Noor, R., and Karbbasi, A.R., "Predicting the Longitudinal Dispersion Coefficient Using Support Vector Machine and Adopted Neuro-Fuzzy Inference System Techniques", Journal of Environmental Engineering Science, Volume 26, pp. 1503-1510.Mehran University Research Journal of Engineering & Technology, Volume 31, No. 1, January, 2012 [ISSN 0254-7821]

144